\documentclass{article} 
\usepackage[final]{colm2026_conference} 
\usepackage[labelfont=bf]{caption}
\usepackage{microtype}
\usepackage{hyperref}
\usepackage{url}
\usepackage{booktabs}
\usepackage{amsmath}
\usepackage{graphicx}
\usepackage{wrapfig}
\usepackage{multirow}
\usepackage{makecell}
\usepackage{cleveref}
\usepackage{subcaption}
\usepackage{siunitx}
\usepackage{array}

\usepackage{lineno}

\definecolor{darkblue}{rgb}{0, 0, 0.5}
\hypersetup{colorlinks=true, citecolor=darkblue, linkcolor=darkblue, urlcolor=darkblue}
\DeclareSIUnit{\million}{M}

\title{Hyperloop Transformers}

\author{Abbas Zeitoun \hspace{3mm} Lucas Torroba-Hennigen \hspace{3mm}  Yoon Kim \vspace{2mm} \\
\vspace{2mm}
Massachusetts Institute of Technology \\
\texttt{\{zeitoun,lucastor,yoonkim\}@mit.edu} \\
\And
}

\newcommand{\subtle}[1]{{\scriptsize\color[gray]{0.4}[#1]}}

\begin{document}

\ifcolmsubmission
\linenumbers
\fi

\maketitle

\begin{abstract}
\vspace{-2mm}
LLM architecture research generally aims to maximize model quality subject to fixed  compute/latency budgets. However, many applications of interest such as edge and on-device deployment are further constrained by the model's memory footprint, thus motivating \emph{parameter-efficient} architectures for language modeling. This paper describes a simple architecture that  improves the parameter-efficiency of LLMs. Our architecture makes use of looped Transformers as a core primitive, which reuse Transformer layers across depth and are thus more parameter-efficient than ordinary (depth-matched) Transformers. We organize the looped Transformer into three blocks---begin, middle, and end blocks---where each block itself consists of multiple Transformer layers, and only the middle block is  applied recurrently across depth. We  augment the looped middle block with \emph{hyper-connections} \citep{mHC}, which expand the   residual stream into matrix-valued residual streams. Hyper-connections are applied only after each loop, and therefore add minimal new parameters and compute cost. Across various model scales, we find that our \emph{Hyper-Connected Looped Transformer (Hyperloop Transformer)} is able to perform well compared to depth-matched Transformer and mHC Transformer baselines despite using approximately 50\% fewer parameters. This performance  persists  through post-training weight quantization, thus positioning Hyperloop Transformers as an attractive architecture for memory-efficient language modeling. 
\vspace{-2mm}
\end{abstract}

\vspace{-2mm}
\section{Introduction}
\vspace{-2mm}
Pushing the Pareto frontier of performance and efficiency is a major goal of modern LLM architecture research. In cloud deployment, efficiency is  measured primarily by latency, which depends on both computation and data movement through the memory hierarchy. Because memory is relatively abundant in such environments, a model's memory footprint is often a secondary concern relative to compute and data movement. This makes parameter-\emph{in}efficient architectures such as mixture-of-experts \citep[MoE;][]{shazeer2017outrageously} viable  for cloud deployment. In contrast, edge and on-device deployments are often constrained not only by compute, but also by the total amount of available memory, which is often orders of magnitude smaller. For example, modern smartphones typically have 8GB--16GB of RAM.
 In such settings, a model's memory footprint becomes a major bottleneck, since it directly affects whether a model can be stored and executed at all.  Even in cloud deployment, fitting a model on fewer accelerators can reduce communication overhead and simplify serving. Looking ahead, frontier models may become large enough that  total parameter memory becomes a first-class constraint even in data-center settings. These factors motivate the study of \emph{parameter-efficient architectures} for  language modeling, where the goal is to push the performance-memory frontier for a given compute constraint.

\emph{Looped Transformers}\footnote{Other terminology used to describe looped Transformers include  \emph{universal Transformers} \citep{dehghani2018universal,tan2023sparse}, \emph{recursive Transformers} \citep{bae2024relaxed,bae2025mixture} , and \emph{recurrent-depth Transformers} \citep{geiping2025scaling,pappone2025two}.} are Transformers that share  parameters across depth, and thus enable greater parameter-efficiency than ordinary Transformers. When the number of loops is variable, they have also been shown to overcome certain theoretical limitations of fixed-depth Transformers \citep{giannou2023looped,yang2023looped,xu2024expressive}, and recent empirical work suggests that they can perform particularly well on some real-world reasoning tasks \citep{geiping2025scaling,zhu2025scaling}. However, when matched for depth, looped Transformers still generally underperform unlooped baselines especially from a perplexity standpoint \citep{saunshi2025reasoning}.

This paper develops a simple looped architecture that outperforms depth-matched Transformer baselines while using approximately half the parameters. Following prior work \citep{bae2025mixture}, we adopt a ``middle cycle'' strategy where we organize the Transformer into begin, middle, and end blocks, and only loop the middle block. We then incorporate a variant of \emph{hyper-connections} \citep{HC,mHC}, which expand the residual stream into multiple streams, into (only) the looped block. Specifically, we apply hyper-connections at the loop level (i.e., only after each loop iteration) instead of at the layer-level, thus incurring minimal additional parameters and compute. We find that our \emph{Hyper-Connected Looped Transformer (Hyperloop Transformer)}  improves the performance-parameter frontier, performing well compared to depth-matched ordinary Transformers with 240M, 1B, and 2B parameters, despite using 50\% fewer parameters. The performance persists through post-training quantization of the model's weights, thus positioning Hyperloop Transformers as an attractive alternative to ordinary Transformers for memory-efficient language modeling.

\vspace{-2mm}
\section{Background}
\vspace{-2mm}
\subsection{Looped Transformers}
\vspace{-2mm}
For a length $T$ input, a Transformer transforms input representations at layer $\mathbf{X}^{(l)} \in \mathbb{R}^{T \times C}$ to obtain the output $\mathbf{X}^{(l+1)} \in \mathbb{R}^{T \times C}$ through an attention layer followed by an MLP layer,
\begin{align*}
    &\mathbf{H}^{(l)} =\text{Attention}(\mathbf{X}^{(l)}; \theta_{\text{attn}}^{(l)}) + \mathbf{X}^{(l)}, && \mathbf{X}^{(l+1)} = \text{MLP}(\mathbf{X}^{(l)}; \theta_{\text{mlp}}^{(l)}) + \mathbf{H}^{(l)}.
\end{align*}
Here $\theta^{(l)}_\text{attn}, \theta^{(l)}_\text{MLP}$ are the layer-specific parameters for multiheaded attention and the feedforward layers respectively.\footnote{The LayerNorm parameters are absorbed into the attention/MLP layers.} Letting $\mathcal{F}_l(\cdot)$  be the application of a Transformer layer $l$, an $L$-layer Transformer then obtains the final output via $\mathbf{X}^{(L)} = \mathcal{F}_L( \dots \mathcal{F}_2(\mathcal{F}_1(\mathbf{X}^{(1)})) \dots)$. Looped Transformers share parameters across depth, e.g., a fully looped model would have  $\mathbf{X}^{(L)} = \mathcal{F}_1( \dots \mathcal{F}_1(\mathcal{F}_1(\mathbf{X}^{(1)})) \dots)$. More recent works have shown that a ``middle cycle'' strategy, which partitions the Transformer layers into beginning, middle, and end blocks\footnote{The begin/end layers are also called prelude/coda or encoder/decoder blocks in the literature.} and only loops the middle block, is particularly effective \citep{bae2025mixture,saunshi2025reasoning}. We also adopt this middle cycle strategy in our architecture.

\vspace{-2mm}
\subsection{Hyper-Connected Transformers}
\vspace{-2mm}
As shown above, each layer of a Transformer  adds to the $C$-dimensional \emph{residual stream}. Hyper-connected Transformers \citep{HC} expand the residual stream to an $n \times C$ dimensional matrix through ``hyper-connections''. In the more recent \emph{manifold-constrained hyper-connections} \citep[mHC; ][]{mHC}, the residual stream at time step $t$ at  depth $l$ (given by $\mathbf{x}_t^{(l)} \in \mathbb{R}^{C}$) is expanded by an expansion factor $n$ to yield $n$ parallel residual streams $\mathbf{y}_t^{(l)} \in \mathbb{R}^{n\times C}$. This expanded residual stream is then read from, written to, and mixed using input-dependent  projections $\mathbf{H}_{l,t}^{\text{pre}}$, $\mathbf{H}_{l,t}^{\text{post}}$, and $\mathbf{H}_{l,t}^{\text{res}}$. Specifically, 
the  transformations at depth $l$ can be computed as follows: \\
\begin{align*}
    &{\mathbf{z}}_{t}^{(l)} = \operatorname{RMSNorm}(\text{flatten}({\mathbf{y}}_t^{(l)})), \\
    & {\mathbf{H}}_{l,t}^{\text{pre}} = \sigma(\alpha_l^{\text{pre}} \cdot (\mathbf{W}_l^{\text{pre}} {\mathbf{z}}_t^{(l)} ) + \mathbf{b}_l^{\text{pre}}), \\
    &{\mathbf{H}}_{l,t}^{\text{post}} = 2\cdot \sigma(\alpha_l^{\text{post}} \cdot (\mathbf{W}_l^{\text{post}} \mathbf{z}_t^{(l)} ) + \mathbf{b}_l^{\text{post}}),  \\
    &{\mathbf{H}}_{l,t}^{\text{res}} = \text{sinkhorn}(\alpha_l^{\text{res}} \cdot \text{reshape}(\mathbf{W}_l^{\text{res}} {\mathbf{z}}_t^{(l)} ) + \mathbf{b}_l^{\text{res}}).
\end{align*}

Here $\mathbf{W}_l^{\text{pre}} \in \mathbb{R}^{n \times nC}$, $\mathbf{W}_l^{\text{post}} \in \mathbb{R}^{n \times nC}$, $\mathbf{W}_l^{\text{res}} \in \mathbb{R}^{n^2 \times nC}$ are linear projections, $\alpha_l^{\text{pre}}$, $\alpha_l^{\text{post}}$,  $\alpha_l^{\text{res}} \in \mathbb{R}$ are learned scalars, $\mathbf{b}_l^{\text{pre}} \in \mathbb{R}^{n}$, $\mathbf{b}_l^{\text{post}} \in \mathbb{R}^{n}$,  $\mathbf{b}_l^{\text{res}} \in \mathbb{R}^{n \times n}$ are learned biases, and  $\text{reshape}(\cdot)$ is an operator that converts an $n^2$-dimensional vector to an $n \times n$ matrix. Finally, $\text{sinkhorn}(\cdot)$ applies the  Sinkhorn-Knopp algorithm, which exponentiates the input and iteratively performs column- and row-normalization,  ensuring that $\mathbf{H}_{l,t}^{\text{res}}$ is doubly stochastic (i.e., on the Birkhoff polytope) in the limit.  \citet{mHC} find that 20 Sinkhorn-Knopp iterations are sufficient.\\

Given the input-dependent matrices  $\mathbf{H}_{l,t}^{\text{pre}} \in \mathbb{R}^{1 \times n}, \mathbf{H}_{l,t}^{\text{post}}  \in \mathbb{R}^{n \times 1}$, $\mathbf{H}_{l,t}^{\text{res}} \in \mathbb{R}^{n \times n}$ and a sub-layer $\mathcal{F}_l \in \{\text{Attention}_l, \text{MLP}_l\}$ of a Transformer layer, mHC applies attention/MLP layers in a smaller residual stream of dimension $C$ via,\footnote{In practice mHC uses different input-dependent matrices for attention and MLP layers.}
\begin{align*}
 \mathbf{y}^{(l+1)}_t = \mathbf{H}_{l,t}^{\text{res}}\mathbf{y}^{(l)}_t  + \mathbf{H}_{l,t}^{\text{post}} \mathcal{F}_l(\mathbf{H}_{l,t}^{\text{pre}} \mathbf{y}_{t}^{(l)}).
\end{align*}
Thus, mHC Transformers make it possible to work with a larger matrix-valued residual stream without incurring much additional compute (since the compute-heavy attention/MLP layers still work with $C$-dimensional inputs/outputs).

\begin{figure}[t]
    \vspace{-4mm}
    \begin{center}
        \includegraphics[width=0.95\linewidth]{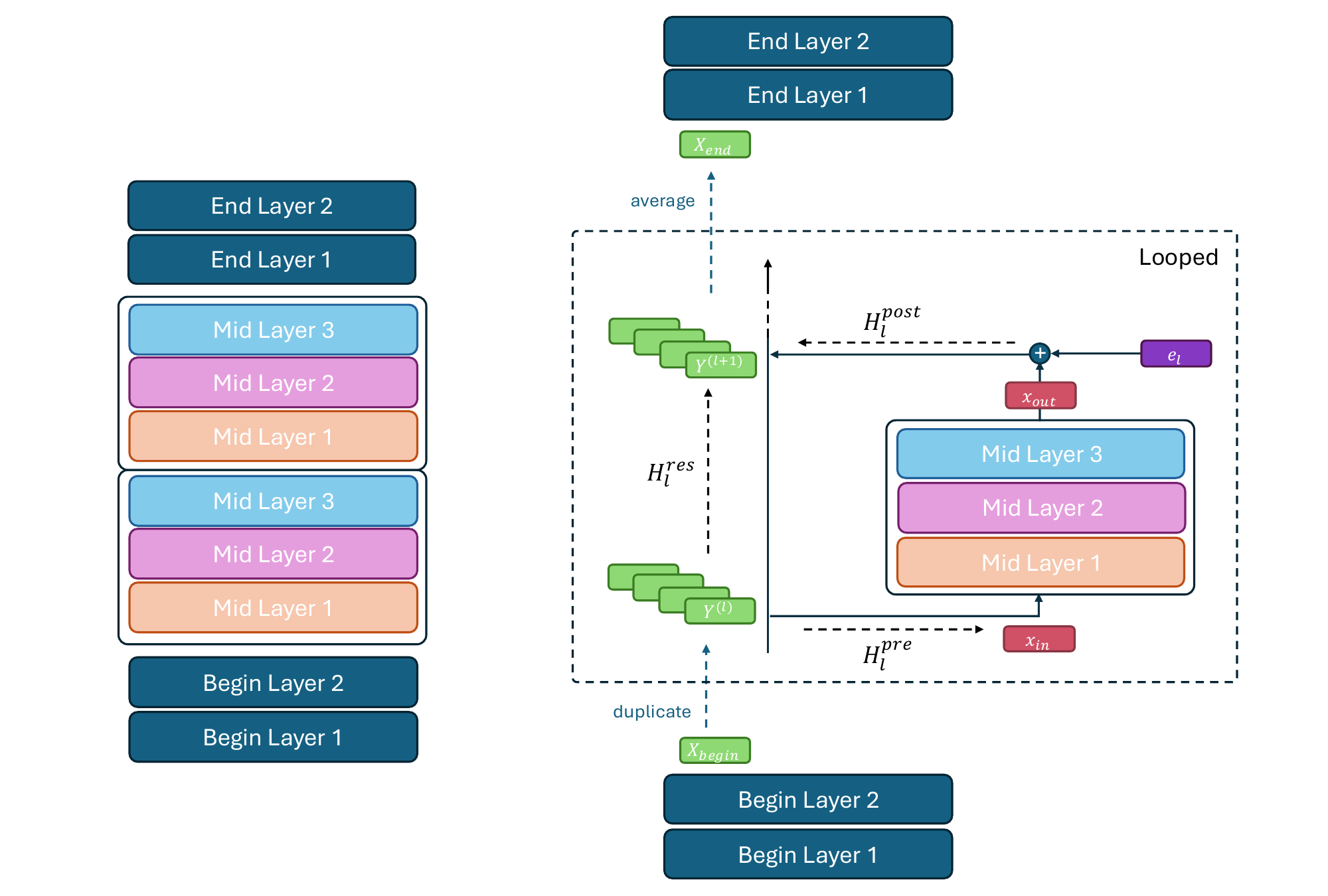}
    \end{center}
    \vspace{-2mm}
\caption{(Left) A vanilla middle-cycle looped  Transformer architecture with two loops. (Right) A Hyperloop Transformer, which uses parallel residual streams that are written to after each loop using hyper-connections \citep{mHC}.}
\vspace{-2mm}
\label{fig:arch}
\end{figure}

\vspace{-2mm}
\section{Hyperloop Transformers}
\vspace{-2mm}
Our architecture, shown in \Cref{fig:arch}, is extremely simple. We partition the Transformer into begin, middle, and end blocks, and then apply (a modification of) hyper-connections at the loop-level when we loop the middle block.

Concretely, let $\mathbf{X}_{\text{begin}}  \in \mathbb{R}^{T \times C}$ be the residual stream after applying the begin block. We expand this to $n$ parallel streams by simply copying it $n$ times, thus giving $\mathbf{Y}^{(0)} \in \mathbb{R}^{T \times n \times C}$, which will serve as input to the hyper-connected looped block. We then compute the input-dependent matrices  $\mathbf{H}_{0,t}^{\text{pre}}, \mathbf{H}_{0,t}^{\text{post}}, \mathbf{H}_{0,t}^{\text{res}} \in \mathbb{R}^{n \times n}$ for all $\{\mathbf{y}_t^{(0)}\}_{t=1}^T$ as above, but using a simpler parameterization of $ \mathbf{H}_{0,t}^{\text{res}}$ given by,
\begin{align*}
    {\mathbf{H}}_{0,t}^{\text{res}} = \text{diag}(\sigma(\alpha_0^{\text{res}} \cdot (\mathbf{W}_0^{\text{res}} {\mathbf{z}}_t^{(0)} ) + \mathbf{b}_0^{\text{res}})),
\end{align*}
where $\mathbf{W}_0^{\text{res}}$ is now an $n \times nC$ matrix (instead of $n^2 \times nC$) and $\mathbf{b}_0^{\text{res}} \in \mathbb{R}^{n}$. 

We  use $\{\mathbf{H}_{0,t}^{\text{pre}}\}_{t=1}^T$ on $\mathbf{Y}^{(0)}$ to obtain the $C$-dimensional input to the middle block, apply the middle block, and then use $\{\mathbf{H}_{0,t}^{\text{post}}\}_{t=1}^T$ to project out into the $n \times C$ residual stream.   We add a ``loop position embedding'' $\mathbf{e}_{l} \in \mathbb{R}^{C}$ after the middle block, resulting in the  recurrence,
\begin{align*}
 \mathbf{y}^{(l+1)}_t = \mathbf{H}_{l,t}^{\text{res}}\mathbf{y}^{(l)}_t  + \mathbf{H}_{l,t}^{\text{post}} \left(\mathcal{F}(\mathbf{H}_{l,t}^{\text{pre}} \mathbf{y}_{t}^{(l)}) +  \mathbf{e}_{l}\right).
\end{align*}
 This process continues for $L$ loops to obtain $\mathbf{Y}^{(L)}$. Finally we average $\mathbf{Y}^{(L)}$ across the parallel streams to obtain $\mathbf{X}_{\text{end}} \in \mathbb{R}^{T \times C}$, which is used as input to the end block.

 Our approach  differs from the original mHC in that (1) we use a simpler  parameterization of $\mathbf{H}_{l,t}^{\text{res}}$ that substitutes the $\text{sinkhorn}(\cdot)$ operator over a dense matrix with a sigmoid over a diagonal matrix (which we found to be sufficient performance-wise while being more efficient), (2) we  add a loop position embedding, which, when viewing the architecture as a ``depth-wise RNN'' with matrix-valued hidden states $\mathbf{Y}^{(0)}$, acts as the input at each time (i.e., loop) step, and (3) we only apply hyper-connections at the loop level, instead of after every attention/MLP layer (so an architecture with 3 loops would have 3 hyper-connections).
Our architecture can also be seen as a more flexible parameterization of looped Transformers, which allows parameters to vary slightly across loop iterations. Concretely, we have loop-specific parameters $\{\mathbf{W}_{l}^{\tau}, \mathbf{b}_{l}^{\tau}, \alpha_l^{\tau}, \mathbf{e}_l\}$ for $\tau \in \{\text{pre}, \text{post}, \text{res}\}$ that can vary across loop iterations $l$. While the number of additional parameters here is still minimal, we posit that this   parameterization allows model representations to change in a more flexible manner compared to ordinary looped Transformers which strictly enforce parameter sharing across each loop iteration.

\vspace{-2mm}
\section{Empirical Study}
\vspace{-2mm}
\subsection{Experimental Setup}
\vspace{-2mm}
We train Hyperloop Transformers  at various scales along with depth-matched vanilla, looped, and mHC Transformer baselines on the FineWeb-Edu dataset \citep{fineweb-edu}. All models make use of SwiGLU MLP layers \citep{shazeer2020glu} and  RoPE  embeddings \citep{su2024roformer}. We use 4 parallel residual streams for both the mHC and  Hyperloop Transformers. For looped models, we allocate (roughly) $25\%$ of the available parameters to the begin block, $25\%$ of the parameters to the end block, and the remaining $50\%$ to the middle block, which is looped three times. This results in looped models that contain half as many parameters as their depth-matched baselines. We ablate on these choices in our ablation study. 

We train  models on $2.5\times$ the Chinchilla-optimal token count of the vanilla Transformer corresponding to their size \citep{hoffmann2022training}. We use the Llama-2 tokenizer to tokenize our data and AdamW as our optimizer, with a linear warmup and cosine decay learning rate schedule. Our full hyperparameters can be found in Appendix~\ref{app:hyperparams}.

\vspace{-2mm}
\subsection{Main Results}
\vspace{-2mm}
For perplexity we evaluate our models on a held-out  set consisting of 50M tokens from the FineWeb-Edu dataset. These are shown in  Table~\ref{main-results}. Our results show that while vanilla Looped Transformers can underperform depth-matched Transformer baselines, the Hyperloop Transformer only needs 150-300K extra parameters (compared to the vanilla Looped Transformer) to  outperform both looped and non-looped depth-matched baseline models. 

While perplexity provides a more robust measure of performance at this scale, we also evaluate our models on downstream tasks. Specifically, we evaluate our models on ARC \citep{clark2018thinksolvedquestionanswering}, COPA \citep{gordon-etal-2012-semeval}, HellaSwag \citep{zellers-etal-2019-hellaswag}, LAMBADA \citep{paperno-etal-2016-lambada}, OpenBookQA \citep{OpenBookQA2018}, PIQA \citep{Bisk2020}, RACE \citep{lai-etal-2017-race}, SciQ \citep{Welbl2017CrowdsourcingMC}, and WinoGrande \citep{sakaguchi2019winogrande}. Interestingly, we find that the looped Transformer also outperforms the vanilla Transformer on most tasks, despite using $50\%$ fewer parameters and despite underperforming the Transformer model in perplexity terms. This outperformance corroborates similar findings reported in the literature \citep{saunshi2025reasoning}. Hyperloop  Transformer outperforms all other baselines overall. Results broken down by task can be found in Appendix~\ref{app:downstream_evals}.

\begin{table}[]
\vspace{-6mm}
    \begin{center}
    \small
    \resizebox{\textwidth}{!}{%
    \begin{tabular}{@{}ll cc S[table-format=4.1,round-precision=1] @{\,}l c c c c@{}}
    \toprule
    \multicolumn{1}{c}{Model}   & {Dim}                                                       & \makecell{Unrolled\\Depth}                   & \makecell{Train\\Tokens}                               & \multicolumn{2}{c}{Params}  &  \makecell{PPL \\ (\texttt{BF16})} &   \makecell{PPL \\ (\texttt{INT4})}                 & \makecell{Task \\ Acc}   & \makecell{Training \\ Toks/s}                   \\ \midrule
    Transformer                                          &  \multirow{4}{*}{1024}        & \multirow{4}{*}{16}       & \multirow{4}{*}{12.5B}         & 238 & M   & 14.65              &   14.85 &   41.1\% & 786K               \\
    mHC                                                 &                                  &    &                            & 241  & M  & 14.55      & 14.73          &  41.1\%   & 514K               \\
    Looped  \subtle{2\texttt{L} $\rightarrow$ 4\texttt{L} $(\times 3)$  $\rightarrow$ 2\texttt{L}}    &         &                         &                                & 135.5  & M& 14.85        & 15.18   & 41.4\%        & 786K                \\
    Hyperloop  \subtle{2\texttt{L} $\rightarrow$ 4\texttt{L} $(\times 3)$  $\rightarrow$ 2\texttt{L}} &     &                             &                                & 135.7  & M& \bfseries  14.40     &\bfseries   14.68&  \bfseries   41.6\%  & 750K                 \\ \midrule
    Transformer                                             &  \multirow{4}{*}{2048}             & \multirow{4}{*}{18}       & \multirow{4}{*}{50B}           & 990.5  & M & 10.19            &  10.27 &   48.0\%    & 367K                \\
    mHC                                                 &                                  &         &                       & 997.5  &M & 10.07  & 10.16               &   48.6\% & 237K                \\
    Looped  \subtle{3\texttt{L} $\rightarrow$ 4\texttt{L} $(\times 3)$  $\rightarrow$ 3\texttt{L}}    &   &                               &                                & 579.4 & M  & 10.02  & 10.24           &  49.2\%      & 367K                 \\
    Hyperloop  \subtle{3\texttt{L} $\rightarrow$ 4\texttt{L} $(\times 3)$  $\rightarrow$ 3\texttt{L}} &       &                           &                                & 579.7  & M &  \textbf{9.65}    & \bfseries 9.81   & \bfseries  49.8\%      & 354K                  \\ \midrule
    Transformer                                              &  \multirow{4}{*}{2048}          & \multirow{4}{*}{38}       & \multirow{4}{*}{100B}          & 2018  & M & 8.60     & 8.71          &  52.8\%     & 181K                  \\
    mHC                      &                           &                                  &                                & 2033 & M & 8.57    & 8.62               &  53.7\% & 109K                 \\
    Looped  \subtle{4\texttt{L} $\rightarrow$ 10\texttt{L} $(\times 3)$  $\rightarrow$ 4\texttt{L}}  &    &                              &                                & 990.5 & M  & 8.68          & 8.97       &  53.3\%    & 183K                 \\
    Hyperloop  \subtle{4\texttt{L} $\rightarrow$ 10\texttt{L} $(\times 3)$  $\rightarrow$ 4\texttt{L}} &    &                              &                                & 990.8 & M & \bfseries  8.49       &  \bfseries   8.59 & \bfseries  54.6\%  & 180K                 \\ \bottomrule
    \end{tabular}
    }
    \end{center}
    \vspace{-2mm}
    \caption{Main results of our architecture and baselines pretrained on FineWeb-Edu. For looped models, [2\texttt{L} $\rightarrow$ 4\texttt{L} $(\times 3)$  $\rightarrow$ 2\texttt{L}] means we have 2 begin layers, 4 middle layers looped 3 times, and 2 end layers. Perplexities are computed with both \texttt{BF16} and \texttt{INT4}, where we use GPTQ to quantize to \texttt{INT4}. Task accuracies are based on \texttt{BF16}. Training throughput measures tokens/second and is based on eight H100s with NVLink.}
    \label{main-results}
    \vspace{-2mm}
\end{table}

\vspace{-2mm}
\paragraph{Post-training quantization.} Post-training quantization of a model's weights is a standard approach for reducing a model's memory footprint. While looped models are \emph{parameter}-efficient, models that are harder to quantize would be practically \emph{memory}-inefficient. Insofar as  models trained with more tokens have been shown to be generally harder to quantize  \citep{huang2024empirical,ouyang2024low}, it is possible that looped models would also be harder to quantize because the looped layers are trained on ``more'' inputs. The interaction effect between looping and quantization has not been investigated before to the best of our knowledge. We quantize our models (originally trained in mixed precision with \texttt{BF16} weights) using GPTQ \citep{frantar-gptq} to \texttt{INT4}, where we modify the GPTQ algorithm so that the Hessian estimation for a looped layer aggregates activations across all inputs to that layer across loops. We use a calibration set of 1024 sequences from FineWeb-Edu, and use a group size of 128 for all model sizes. The  perplexities of the resulting \texttt{INT4} models are presented in Table~\ref{main-results}. Our results show that while looped Transformers can indeed be somewhat more sensitive to lower-precision quantization compared to non-looped models, hyper-connections help alleviate some of the performance degradation resulting from quantization. As a result, Hyperloop Transformers continue to perform well in the weight-only quantization setting.

\paragraph{Training efficiency.} Do the extra hyper-connections add training overhead? We measure the training throughput of each of the pretrained models on a single node with $8\times\text{H100}$ GPUs with NVLink and present the results in Table~\ref{main-results}. The models used for these measurements were implemented in PyTorch and compiled with \texttt{torch.compile} but without any further optimizations. Our results show that a straightforward PyTorch implementation of our approach only incurs minimal slowdown compared to the Transformer and Looped Transformer baselines. This can be attributed to only applying hyper-connections at the \emph{inter-loop} level, and also using a simpler structure for $\mathbf{H}^{\text{res}}$ resulting in very little memory and compute overhead.
On the other hand, a straightforward  implementation of the mHC Transformer results in nontrivial slowdowns.  This overhead can be brought down in theory with the proper low-level optimizations---for example, \citet{mHC} report a 6.7\% overhead with their specialized training kernel. However, this kernel is not publicly available to the best of our knowledge, and the fact that our approach adds almost no overhead without requiring any sophisticated systems engineering is an additional benefit.

\begin{wraptable}{r}{0.5\textwidth}
   \centering
    \small
    \begin{tabular}{@{}lccc@{}}
    \toprule
    & & \multicolumn{2}{c}{Train Tokens}  \\
    \multicolumn{1}{c}{Model} & Params   & 12.5B & 100B    \\ \midrule
    Transformer  & 238.0 M &14.65 &  12.15 \\
    mHC            & 241.0 M   & 14.55 & 12.16 \\
    Looped            & 135.5 M   & 14.85 &  12.56 \\
    Hyperloop       & 135.7 M   & 14.40 &  12.19 \\
    \bottomrule
    \end{tabular}
    \caption{Perplexity results for our smallest (16 layer) models trained on more tokens.}
    \vspace{-2mm}
    \label{tab:overtrained_results}
\end{wraptable}

\paragraph{Training for more tokens.} The number of tokens in our training set is 50$\times$  the number of parameters of the non-looped baseline models, i.e., 2.5$\times$ that of the compute-optimal 20$\times$ recipe suggested by \citet{hoffmann2022training}. However, modern models are typically trained for many more tokens than is Chinchilla-optimal. For example, LLaMA3-8B \citep{grattafiori2024llama} was trained on 15T tokens, while OLMo3-7B \citep{olmo2025olmo} was trained on 6T tokens. Would the benefits of looped models diminish in such overtraining regimes? To investigate this, we train our smallest class of models, corresponding to 240M non-looped parameters, on 100B tokens from FineWeb-Edu. This corresponds to 20$\times$ the Chinchilla-optimal number of tokens, or approximately 400 training tokens per parameter. The results are presented in Table~\ref{tab:overtrained_results}. We find that while looped Transformers underperform non-looped baselines, Hyperloop Transformers remain competitive with these baselines despite being in an overtrained setting.

\vspace{-2mm}
\subsection{Ablations}
\begin{figure}[t]
\vspace{-8mm}
\centering
\includegraphics[width=0.9\linewidth]{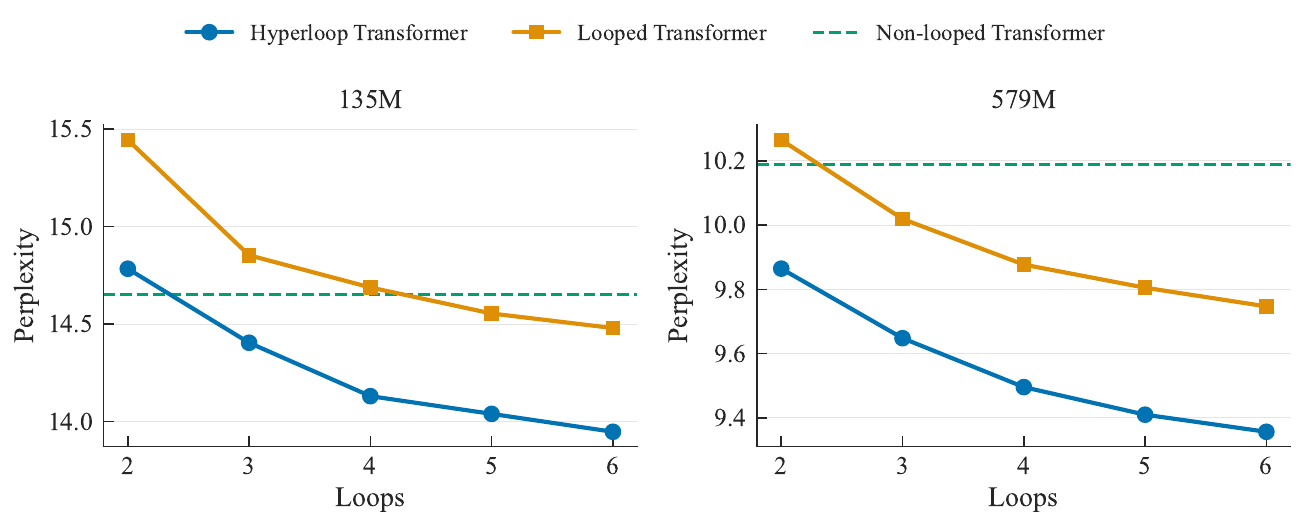}
\vspace{-3mm}
\caption{Perplexity numbers as the number of loops is varied for the 135M (left) and 579M (right) parameter looped models. The non-looped Transformer baselines have 238M (left) and 991M (right) parameters. Each loop consists of 4 Transformer layers.}
\vspace{-4mm}
\label{fig:ppl_loops}
\end{figure}

\vspace{-2mm}
\paragraph{Number of loops.} Our main experiments use 3 loops. How does performance vary as a function of the number of loops, if we hold the parameters constant? We  vary the number of loops from 2 to 6  for the 136M- and 579M-parameter looped and Hyperloop Transformers, and show the results in \Cref{fig:ppl_loops}. We observe  diminishing returns as we increase the number of loops, although Hyperloop Transformer dominates the Looped Transformer in all cases.

The above experiments hold the parameter count constant and vary the depth by changing the number of loops. We next experiment with restructuring the middle loop by making it smaller but looping it more to keep the depth constant. The results are shown in Table~\ref{tab:loop-param}. We find that we can obtain even greater parameter-efficiency at the cost of a small performance hit. For example, our 477M parameter model still outperforms the full 1B Transformer. 

We now conduct more ablations focusing on the 136M-parameter/12.5B-token setting.

\begin{table}[t]
\small
\center
\vspace{-6mm}
\begin{tabular}{@{}lccc S[detect-all, table-format=2.3,round-precision=2] } \toprule
                    \multicolumn{1}{c}{Model} &Structure           & Unrolled Depth             & Params & \multicolumn{1}{c}{PPL} \\ \midrule
&2\texttt{L} $\rightarrow$ 4\texttt{L} $(\times 3)$ $\rightarrow$ 2\texttt{L}     & \multirow{3}{*}{16 layers} & 136 M       & 14.853                 \\
Looped Transformer &2\texttt{L} $\rightarrow$ 3\texttt{L} $(\times 4)$ $\rightarrow$ 2\texttt{L}   &                            & 123 M       & 15.184                  \\
& 2\texttt{L} $\rightarrow$  2\texttt{L} $(\times 6)$ $\rightarrow$ 2\texttt{L}      &                            & 110 M       & 15.763                 \\ \midrule
& 2\texttt{L} $\rightarrow$ 4\texttt{L} $(\times 3)$ $\rightarrow$ 2\texttt{L}   & \multirow{3}{*}{16 layers} & 136 M       & 14.404                  \\
Hyperloop Transformer & 2\texttt{L} $\rightarrow$ 3\texttt{L} $(\times 4)$ $\rightarrow$ 2\texttt{L} &                            & 123 M       & 14.618                 \\
& 2\texttt{L} $\rightarrow$  2\texttt{L} $(\times 6)$  $\rightarrow$ 2\texttt{L}  &                            & 110 M       & 15.056                 \\ \midrule
& 3\texttt{L} $\rightarrow$ 4\texttt{L} $(\times 3)$  $\rightarrow$ 3\texttt{L}     & \multirow{3}{*}{18 layers} & 579 M       & 10.019                 \\
Looped Transformer & 3\texttt{L} $\rightarrow$  3\texttt{L}  $(\times 4)$ $\rightarrow$ 3\texttt{L}    &                            & 528 M       & 10.124                 \\
& 3\texttt{L} $\rightarrow$  2\texttt{L} $(\times 6)$  $\rightarrow$ 3\texttt{L}      &                            & 477 M       & 10.357                 \\ \midrule
& 3\texttt{L} $\rightarrow$ 4\texttt{L} $(\times 3)$ $\rightarrow$ 2\texttt{L}   & \multirow{3}{*}{18 layers} & 579 M       & 9.648                   \\
Hyperloop Transformer  &3\texttt{L} $\rightarrow$  3\texttt{L} $(\times 4)$  $\rightarrow$ 3\texttt{L} &                            & 528 M       & 9.717                  \\
& 3\texttt{L} $\rightarrow$  2\texttt{L} $(\times 6)$  $\rightarrow$ 3\texttt{L}  &                            & 477 M       & 9.862                 \\ \bottomrule
\end{tabular}
\vspace{-2mm}
\caption{Performance of the looped and Hyperloop Transformers as we vary the looping structure while keeping the depth fixed at 16 or 18 layers. }
\vspace{-5mm}
\label{tab:loop-param}
\end{table}

\begin{wraptable}{r}{0.4\textwidth}
\vspace{-0.4cm}
\small
        \centering
        \begin{tabular}{@{}c S[detect-all, table-format=2.3,round-precision=2]@{}}
        \toprule
          \multirow{2}{*}{\makecell{Number of \\ Streams $n$}} & {\multirow{2}{*}{PPL}} \\
          &                        \\ \midrule
        2 & 14.429\\
        4 & 14.404\\
        6 & 14.379\\
        8 & 14.388 \\
        10 & 14.349\\ \bottomrule
        \end{tabular}
           \vspace{-2mm}
        \caption{Hyperloop Transformer performance as we vary the number of parallel residual streams.}
        \vspace{-5mm}
        \label{num_streams_results}
\end{wraptable}
\vspace{-2mm}
\paragraph{Number of parallel streams.} We pick $n=4$ parallel residual streams as recommended by the original mHC work. Insofar as the number of parallel streams  provides a parameter-efficient axis with which to scale up the model, can we get further gains by increasing $n$? \Cref{num_streams_results} shows the results on a 135M-parameter Hyperloop Transformer with 3 loops and a varying number of parallel residual streams, where we observe diminishing returns on $n$. Thus, while having matrix-valued residual streams does improve performance, this axis of scaling rapidly faces diminishing returns.

\vspace{-2mm}
\paragraph{Number of hyper-connections.} Recall that our Hyperloop Transformer only applies hyper-connections after each loop, which results in minimal additional parameter/compute overhead. We ablate on the number of hyper-connections used within the looped block of a 135M-parameter Hyperloop Transformer. \begin{wraptable}{r}{0.37\textwidth}
\small
        \centering
\begin{tabular}{r @{\;} l l}
         \toprule
  \multicolumn{2}{c}{Hyper}        & {\multirow{2}{*}{PPL}} \\
  \multicolumn{2}{c}{connections} &                        \\ \midrule
  12 & \subtle{every layer}    & 14.45 \\
   6 & \subtle{every 2 layers} & 14.50 \\
   4 & \subtle{every 3 layers} & 14.50 \\
   3 & \subtle{every loop (ours)}     & 14.40 \\
   2 & \subtle{every 6 layers} & 14.50 \\
   1 & \subtle{every 12 layers}& 14.63 \\ \bottomrule
\end{tabular}
                   \vspace{-2mm}
        \caption{Performance as we vary the number of HCs.}
        \vspace{-2mm}
                 \label{num_hc_results}
\end{wraptable} 
To do so, we fix the number of loops and only vary the number of hyper-connections, or equivalently, the number of Transformer blocks skipped over by a single hyper-connection. This results in some hyper-connections being applied within loops or across Transformer blocks from different loops in some setups. 
The sub-layers within a Transformer block retain their own original skip-connections, even when hyper-connections are applied after every block.  \begin{wraptable}{r}{0.35\textwidth}
       \vspace{-4mm}
       \small
        \centering
        \begin{tabular}{@{}lc@{}}
        \toprule
        Parameterization & {PPL} \\ \midrule
        Identity & 14.61 \\
        Sinkhorn & 14.59 \\
        Diagonal \subtle{ours} & 14.40 \\ \bottomrule
        \end{tabular}
                   \vspace{-2mm}
        \caption{Ablations on  parameterization of the transition matrix $\mathbf{H}^{\text{res}}$.}
               \vspace{-3mm}
        \label{res_param_results}
\end{wraptable}
\Cref{num_hc_results} shows that applying hyper-connections after every loop (instead of every layer) is the most performant setup. This is perhaps unintuitive, given that the every-layer setup uses the most compute/parameters. Our results potentially indicate that at least in the looped case, one must be more careful about choosing where to apply hyper-connections.

\vspace{-2mm}
\paragraph{$\mathbf{H}^{\text{res}}$ parameterization.} The Hyperloop Transformer simplifies the mHC formulation by using a simpler diagonal transition matrix $\mathbf{H}^{\text{res}}$ for the parallel residual stream, in contrast to the doubly-stochastic structure used in mHC. This leads to fewer parameters and less compute. Does this potentially hurt the performance of our approach? As shown in Table~\ref{res_param_results}, we find that this is not the case. The simplest identity parameterization  only slightly underperforms the Sinkhorn parameterization, while our (data-dependent) diagonal parameterization further improves performance.
\begin{wraptable}{r}{0.4\textwidth}
       \vspace{-5mm}
   \centering
    \small
    \begin{tabular}{@{}llc@{}}
    \toprule
    \multicolumn{1}{c}{LoRA Rank}    & Params & PPL    \\ \midrule
    0 \subtle{Looped Transformer} & 135.5M & 14.85 \\
    4            & 136.0M   & 14.85 \\
    8            & 137.0M   & 14.81 \\
    16           & 139.0M   & 14.80 \\
    32           & 143.0M   & 14.77 \\
    \midrule
    Transformer & 238.0M & 14.65 \\
    Hyperloop  &136.0M & 14.40 \\
    \bottomrule
    \end{tabular}
        \vspace{-2mm}
    \caption{Experiments on allowing Transformer layers to change across loop iterations with LoRA.}
    \vspace{-10mm}
    \label{tab:lora_results}
\end{wraptable}
\vspace{-2mm}
\paragraph{Comparison vs. LoRA-Looped  Transformers.} Since the weights for the hyper-connections are different across loops, our Hyperloop Transformer allows for the looped block to be slightly different across loops. Would using LoRA to modify the parameters across loops (as in relaxed recursive Transformers; \citealp{bae2024relaxed}) perform better? The results of these experiments are shown in  Table~\ref{tab:lora_results}. We find that allowing parameters to change across loops with LoRA does help slightly, but the  Hyperloop Transformer provides a much more parameter-efficient approach to improving performance.

\vspace{-2mm}
\subsection{Analysis}

To better understand our model's inner workings, we conduct a series of qualitative analyses of its internal representations on 50M tokens from the FineWeb-Edu dataset.

\begin{figure}[t]
\vspace{-4mm}
    \centering
    \includegraphics[width=1\linewidth]{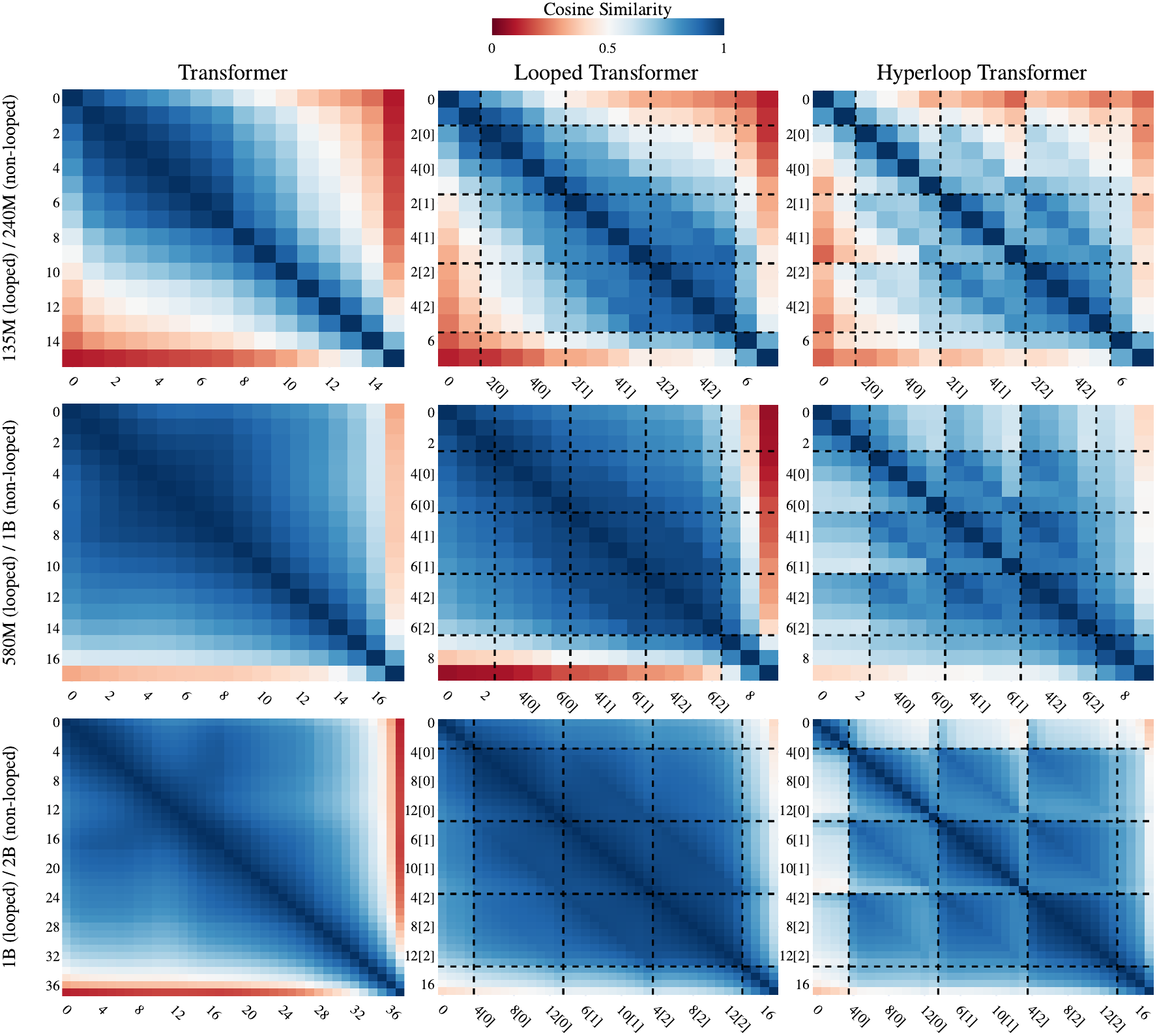}
    \caption{Pairwise cosine similarity between inner residual streams at each (effective) layer, across model scales (rows) and architectures (columns). 
    }
    \label{fig:cosine-sim}
\end{figure}

\begin{wraptable}{r}{0.4\textwidth}
    \centering
    \vspace{-4mm}
    \begin{tabular}{c S[table-format=1.3] S[table-format=1.3]}
    \toprule
    Params & {Looped} & {Hyperloop} \\
    \midrule
    136M & 0.7429 & 0.7382 \\
    579M & 0.9152 & 0.8723 \\
    991M & 0.9226 & 0.8714 \\
    \bottomrule
    \end{tabular}
    \caption{Average cosine similarity between corresponding layers across loop iterations.}
    \label{tab:cross_loop_similarity}
    \vspace{-2mm}
\end{wraptable}
  \vspace{-2mm}
\paragraph{Representation similarity.}
We hypothesize that the outperformance of the Hyperloop architecture is supported in part by hyper-connections allowing for the model representations to be less constrained than in the ordinary looped case. To investigate this, we analyze the cosine similarity of the residual stream as we vary the depth in \Cref{fig:cosine-sim}.
We  see that both looped models exhibit similarity within the looped blocks; in particular, we also see that the representations output by the same layer \emph{across} loops exhibit higher-than-expected similarity.
\Cref{tab:cross_loop_similarity} quantifies the average similarity of layers across loops (e.g., comparing representations of middle layer 1 across loops) for all the looped layers. We find that Hyperloop models' representations are indeed less similar, supporting our hypothesis.

\begin{figure}[t]
    \centering
\vspace{-4mm}
    \includegraphics[width=1\linewidth]{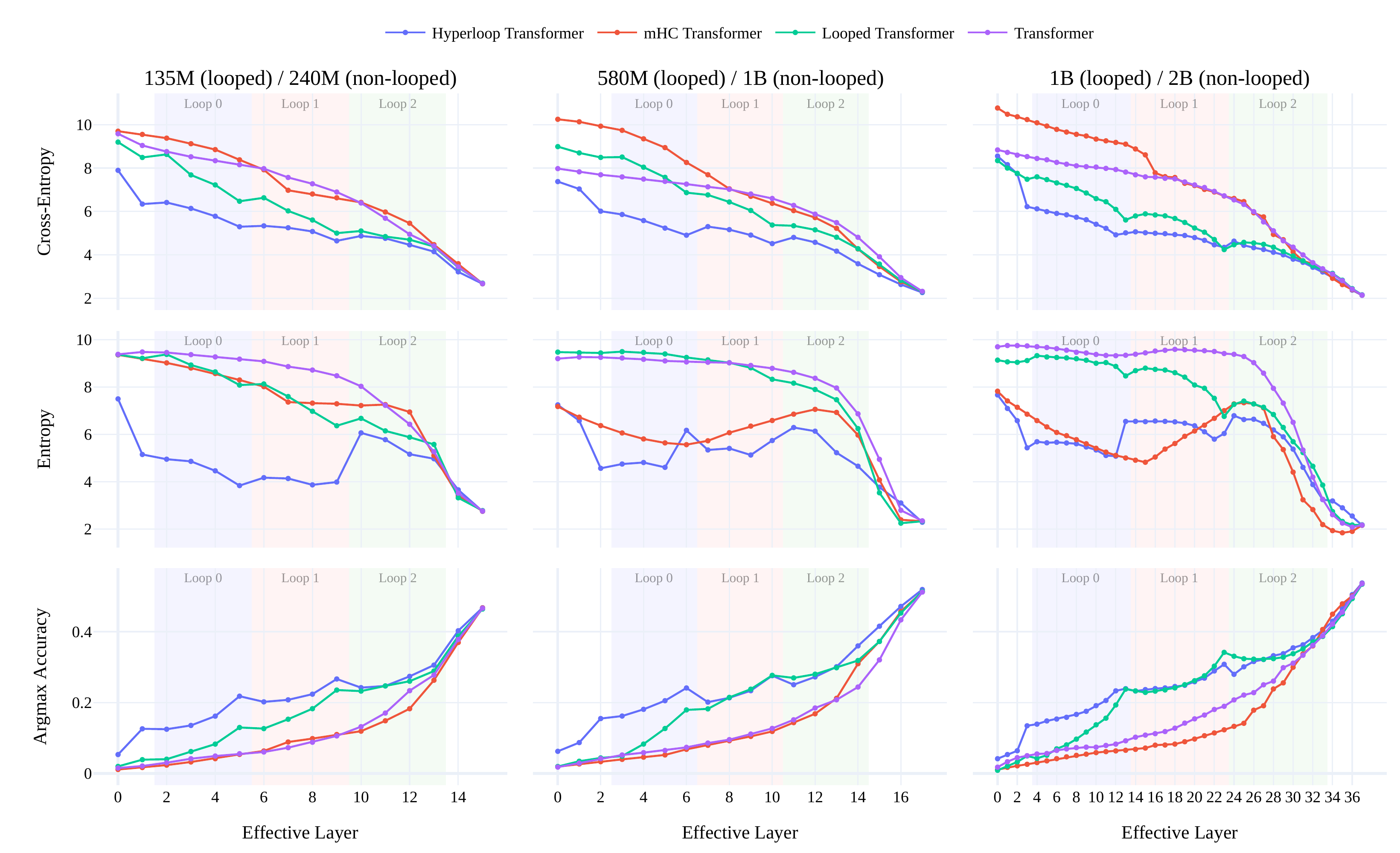}
    \caption{Logit lens-inspired analysis across model scales. Each column corresponds to a model scale, and each row shows a different metric: average cross-entropy (top), average entropy of vocabulary distribution (middle), and greedy decoding accuracy (bottom), computed by mapping the outer residual stream via the language modeling head. Loop boundaries are indicated at the top of each panel, though they only apply to looped models.}
    \label{fig:logit-lens}
\end{figure}

\vspace{-2mm}
\paragraph{Logit lens.}
We also perform a logit lens-style analysis~\citep{logit-lens}.
We observe that the ``outer'' residual streams
(i.e., the parallel streams in the mHC/Hyperloop Transformers; the regular stream in the other Transformers)
are loosely aligned to the vocabulary space.\footnote{For the Hyperloop model, we can compute the effective value of the parallel residual streams by performing an early merge operation from the intra-loop residual stream to the parallel  stream.}
We are thus able to push these representations through the language modeling head to get a distribution over the next token.
From this, we can compute the evolution of the cross-entropy, entropy, and the accuracy of the argmax of the distribution, as shown in \Cref{fig:logit-lens}.
We see that both Hyperloop and the vanilla looped transformer produce representations that are more aligned with the vocabulary distribution, likely as the looping forces the models to operate closer to the vocabulary space.
The fact that both looped models exhibit maximum alignment toward the vocabulary distribution at the end of a loop further corroborates this claim.
Interestingly, our approach produces models with \emph{higher} alignment than the vanilla looped models, suggesting that the hyper-connections offer additional regularization in this direction. This potentially indicates that Hyperloop Transformers could be more amenable to early-exit-style inference strategies and enable compute savings.

\vspace{-2mm}
\section{Discussion}
\vspace{-2mm}
This work shows that combining recent hyper-connections \citep{mHC} with looped Transformers can push the parameter-performance frontier of language models. We study several methods for  incorporating hyper-connections, and show that doing so at the loop-level with a simple data-dependent diagonal transition matrix is  effective, while incurring minimal additional parameters/compute. We find suggestive evidence that the outperformance of our approach is supported in part by the hyper-connections allowing the model representations of looped layers to deviate more flexibly. While we have primarily focused on parameter-efficiency controlling for compute, our logit lens analysis further suggests that Hyperloop Transformers could enable compute efficiency gains through early-exit style inference strategies. 

Our main limitation is scale. While we performed experiments that were reasonable on academic compute, it is unclear as to whether the overall efficiency gain (i.e., Hyperloop Transformers matching Transformers with 50\% fewer parameters) would hold at even larger scales, although we did find that Hyperloop Transformers were effective in the overtrained regime for the smaller models. While the present work was mostly motivated by pushing the performance-efficiency frontier, looped Transformers have been suggested as a better architecture for enabling test-time scaling and improved reasoning \citep{saunshi2025reasoning,geiping2025scaling,zhangmodr,kohli2026loop}. It would thus be interesting to train much deeper Hyperloop Transformers (with the hyper-connection parameters potentially shared across loops to enable generalization to longer loops than seen in training) to investigate its test-time scaling and reasoning capabilities.

\vspace{-2mm}
\section{Related Work}
\vspace{-2mm}

\paragraph{Looped Transformers.} Looped Transformers were first proposed by \citet{dehghani2018universal} and applied to BERT-style models in ALBERT \citep{lan2019albert}. Modern variants of looped Transformers were initially studied in synthetic settings where they were found to generalize better on certain kinds of synthetic tasks \citep{csordas2021devil,csordas2022neural,giannou2023looped,yang2023looped,xu2024expressive}. However, more recent works have shown the empirical effectiveness of looped models on real language modeling.  \citet{csordas2024moeut} generalize universal Transformers to the mixture-of-experts case. 
\citet{saunshi2025reasoning}  find that despite underperforming unlooped baselines from a perplexity standpoint, looped models perform better on certain kinds of reasoning tasks. \citet{kohli2026loop} study synthetic multi-hop reasoning with looped models and find that they can generalize to more hops than seen in training. \citet{bae2024relaxed} and \citet{mcleish2025teachingpretrainedlanguagemodels}  convert pretrained unlooped Transformers into a looped architecture. \citet{bae2025mixture} propose a looped architecture that allocates a dynamic number of loops on a per-token basis. \citet{geiping2025scaling} train a looped model on a variable number of loops and show that the model's performance on downstream tasks scales with the number of loops at test-time. \citet{jeddi2026loopformer} also enable flexible number of loops at inference time through conditioning on depth during training.  \citet{zhu2025scaling} train looped language models through all stages of a modern language modeling pipeline and propose an entropy-regularized objective for early exiting after a dynamic number of loops.  \citet{yu2026spiralformer} show the effectiveness of looping blocks after downscaling the input.  \citet{prairie2026parcae} develop a more stable parameterization of looped Transformers and derive scaling laws as the number of loops is varied, holding parameter count fixed (i.e., as in \Cref{fig:ppl_loops}). \citet{schwethelm2026how}  study scaling laws of looped Transformers by varying the  looping structure and number of parameters.
Finally, \citet{blayeney2026mechanistic}  perform a mechanistic analysis of looped Transformers and find that the latent states follow a cyclic trajectory.

\vspace{-2mm}
\paragraph{Residual connections in Transformers.} Our work is also related to approaches that modify the residual stream connection patterns in Transformers. \citet{HC} and \citet{mHC} expand the residual stream into a {residual matrix}, allowing richer connections from earlier to later layers in the model. \citet{pagliardini2024denseformerenhancinginformationflow} expand the residual stream differently, averaging the hidden states at the output of a transformer block with earlier hidden states using different sparsity patterns. \citet{xiao2025muddformerbreakingresidualbottlenecks}, \citet{heddes2025deepcrossattention}, and \citet{kimiteam2026attentionresiduals} take that a step further, allowing the model to {attend} to previous hidden states along the depth axis. 
We leave the integration of looped Transformers with other residual connection patterns to future work. 

\vspace{-1mm}
\section{Conclusion}
\vspace{-1mm}
We propose a simple architecture that combines hyper-connections with looped Transformers and improves the parameter-efficiency of language models while adding minimal additional compute at training and deployment.
\vspace{-1mm}
\section*{Acknowledgments}
\vspace{-1mm}
 We thank Junhyun Lee, Munjo Kim, and Oliver Sieberling for helpful discussions. This study was supported by a Samsung Research grant and the AI2050 program at Schmidt Sciences (Grant G-25-67980). 

\bibliography{colm2026_conference}
\bibliographystyle{colm2026_conference}

\clearpage
\appendix
\vspace{-2mm}
\section{Hyperparameters}
\vspace{-2mm}
 We use a model dimension of $1024$ for our 240M/136M-parameter models and a model dimension of $2048$ for our larger models. We set our SwiGLU feed-forward dimension to be $2.75\times$ the model dimension across all model sizes. All model sizes use Multi-Head Attention with 16 attention heads and a RoPE base of 10000 for their position embeddings. We use a weight-untied unembedding matrix and include its number of parameters in our reported model sizes.

Our models are trained on batches of 256 sequences of length 2048, corresponding to 524K tokens per batch. Across all training runs, we use a max learning rate of $4 \times 10^{-4}$, with cosine decay down to $4 \times 10^{-5}$. We use 1000 warmup steps for our 240M/136M models and 2000 warmup steps for the larger models. For AdamW, we use $(\beta_1, \beta_2) = (0.9, 0.95)$ and a weight decay of $0.1$. We use gradient normalization with max value $1.0$. 
\label{app:hyperparams}
\vspace{-2mm}
\section{Downstream Task Evaluations}
\vspace{-2mm}
\label{app:downstream_evals}
Table~\ref{tab:downstream_tasks_combined} shows the downstream task results broken down by tasks.

\begin{table*}[t]
\vspace{-2mm}
\centering
\small
\begin{tabular}{@{}llcccc@{}}
\toprule
Model Parameters & Task & Transformer & \makecell{mHC} & \makecell{Looped} & \makecell{Hyperloop} \\
\midrule

\multirow{12}{*}{\makecell{240M (non-looped)\\ / 136M (looped)}} 
& ARC-Challenge      & 19.45\% & \textbf{21.25\%} & 19.71\% & 20.56\% \\
& ARC-Easy           & 49.24\% & 49.79\% & 49.45\% & \textbf{50.63\%} \\
& COPA               & 62.00\% & 60.00\% & 62.00\% & \textbf{63.00\%} \\
& HellaSwag$^\star$  & 31.96\% & 31.87\% & 31.37\% & \textbf{32.00\%} \\
& LAMBADA (OpenAI)   & 24.14\% & 24.76\% & \textbf{25.17\%} & 25.09\% \\
& LAMBADA (Standard) & 17.95\% & 17.66\% & \textbf{18.03\%} & 17.89\% \\
& OpenBookQA$^\star$ & 30.60\% & \textbf{31.40\%} & 30.80\% & 31.20\% \\
& PIQA$^\star$       & 61.53\% & 61.37\% & 60.94\% & \textbf{63.33\%} \\
& RACE               & 29.19\% & 26.32\% & \textbf{29.86\%} & 28.71\% \\
& SciQ               & 76.60\% & \textbf{77.00\%} & 74.50\% & 75.20\% \\
& WinoGrande         & 49.88\% & 50.12\% & \textbf{53.04\%} & 50.20\% \\
& \textit{Average}   & \textit{41.14\%} & \textit{41.05\%} & \textit{41.35\%} & \textit{\textbf{41.62\%}} \\
\midrule
\multirow{12}{*}{\makecell{1B (non-looped)\\ / 579M (looped)}} 
& ARC-Challenge      & 25.68\% & 27.90\% & 27.73\% & \textbf{28.07\%} \\
& ARC-Easy           & 59.30\% & 60.14\% & \textbf{62.46\%} & 62.08\% \\
& COPA               & 70.00\% & 71.00\% & \textbf{73.00\%} & 68.00\% \\
& HellaSwag$^\star$  & 42.07\% & 42.80\% & 43.98\% & \textbf{46.22\%} \\
& LAMBADA (OpenAI)   & 36.64\% & 36.11\% & 37.16\% & \textbf{39.05\%} \\
& LAMBADA (Standard) & 28.20\% & 27.94\% & 28.18\% & \textbf{30.62\%} \\
& OpenBookQA$^\star$ & \textbf{34.40\%} & 33.40\% & 33.40\% & 33.80\% \\
& PIQA$^\star$       & 66.43\% & 67.08\% & 67.14\% & \textbf{68.72\%} \\
& RACE               & 30.53\% & 31.10\% & \textbf{31.77\%} & \textbf{31.77\%} \\
& SciQ               & 82.80\% & 83.90\% & 84.70\% & \textbf{86.30\%} \\
& WinoGrande         & 52.17\% & 52.96\% & 51.62\% & \textbf{53.04\%} \\
& \textit{Average}   & \textit{48.02\%} & \textit{48.58\%} & \textit{49.19\%} & \textit{\textbf{49.79\%}} \\
\midrule
\multirow{12}{*}{\makecell{2B (non-looped)\\ / 1B (looped)}} 
& ARC-Challenge      & 31.66\% & 31.74\% & 32.17\% & \textbf{33.70\%} \\
& ARC-Easy           & 68.18\% & 66.96\% & 68.60\% & \textbf{69.02\%} \\
& COPA               & 70.00\% & \textbf{74.00\%} & 72.00\% & \textbf{74.00\%} \\
& HellaSwag$^\star$  & 51.44\% & 51.87\% & 52.21\% & \textbf{53.93\%} \\
& LAMBADA (OpenAI)   & 42.89\% & 43.94\% & 43.37\% & \textbf{45.55\%} \\
& LAMBADA (Standard) & 35.16\% & 35.44\% & 34.06\% & \textbf{37.96\%} \\
& OpenBookQA$^\star$ & \textbf{37.80\%} & 37.40\% & 36.20\% & 36.20\% \\
& PIQA$^\star$       & \textbf{71.49\%} & 70.89\% & 71.33\% & 71.27\% \\
& RACE               & 32.54\% & \textbf{34.64\%} & 32.34\% & 33.59\% \\
& SciQ               & 85.50\% & 87.60\% & \textbf{88.50\%} & 88.20\% \\
& WinoGrande         & 53.83\% & 56.35\% & 55.41\% & \textbf{57.06\%} \\
& \textit{Average}   & \textit{52.77\%} & \textit{53.71\%} & \textit{53.29\%} & \textit{\textbf{54.59\%}} \\
\bottomrule
\end{tabular}
\vspace{-2mm}
\caption{Accuracies on downstream tasks across three model-size settings. For tasks marked with a $\star$, we normalize the log-likelihood of the different multiple-choice continuations by the number of tokens.}
\vspace{-2mm}
\label{tab:downstream_tasks_combined}
\end{table*}

\vspace{-2mm}
\section{Results with Different Initializations}
\vspace{-2mm}
The results reported in the main body of the paper were obtained using models initialized with Pytorch defaults. We conducted additional experiments experiments on 240M/136M-parameter models with different initializations and learning rates. 

Concretely, we initialize all model weights from a $\mathcal{N}(0, 0.02)$ distribution by default, except for layers that write to the residual stream, which are initialized from a $\mathcal{N}\left(0, 0.02/\sqrt{2* \texttt{total\_unrolled\_depth}}\right)$ as in the GPT-2 paper. We tuned the learning rates across \{4e-4, 8e-4, 1.2e-4, 1.6e-3, 2e-3\} for all models (1.6e-3 generally performed best). The resulting  perplexities of the best settings are reported in Table~\ref{tab:updated_init_results}.

Under this arguably more standard initialization scheme, the mHC Transformer performs particularly well, while the Hyperloop Transformer is able to match but not outperform the baseline  Transformer, although it does significantly improve upon the ordinary Looped Transformer.
\begin{table}[]
\vspace{-6mm}
    \begin{center}
    \small
    \begin{tabular}{@{}ll cc S[table-format=4.1,round-precision=1] @{\,}l c@{}}
    \toprule
    \multicolumn{1}{c}{Model}   & {Dim}                                                       & \makecell{Unrolled\\Depth}                   & \makecell{Train\\Tokens}                               & \multicolumn{2}{c}{Params}  &  \makecell{PPL \\ (\texttt{BF16})}                   \\ \midrule
    Transformer                                          &  \multirow{4}{*}{1024}        & \multirow{4}{*}{16}       & \multirow{4}{*}{12.5B}         & 238 & M   & 12.94               \\
    mHC                                                 &                                  &    &                            & 241  & M  & \bfseries 12.67                \\
    Looped  \subtle{2\texttt{L} $\rightarrow$ 4\texttt{L} $(\times 3)$  $\rightarrow$ 2\texttt{L}}    &         &                         &                                & 135.5  & M& 13.47                 \\
    Hyperloop  \subtle{2\texttt{L} $\rightarrow$ 4\texttt{L} $(\times 3)$  $\rightarrow$ 2\texttt{L}} &     &                             &                                & 135.7  & M& 13.08                  \\
    \bottomrule
    \end{tabular}
    \end{center}
    \vspace{-2mm}
    \caption{Perplexity results of our architecture and baselines pretrained on FineWeb-Edu under further tuned initializations and hyper-parameters.}
    \label{tab:updated_init_results}
    \vspace{-2mm}

\end{table}
\end{document}